\documentclass[australian,british]{llncs}
\usepackage[T1]{fontenc}
\usepackage[latin9]{inputenc}
\usepackage{array}
\usepackage{booktabs}
\usepackage{multirow}
\usepackage{amsmath}
\usepackage{stackrel}
\usepackage{graphicx}

\makeatletter

\providecommand{\tabularnewline}{\\}
\usepackage{microtype}
\usepackage{times}
\usepackage[T1]{fontenc}

\makeatother

\usepackage{babel}
\begin{document}
\title{MP-PINN: A Multi-Phase Physics-Informed Neural Network for Epidemic Forecasting}
\author{Thang Nguyen\inst{1} \and Dung Nguyen\inst{1} \and Kha Pham\inst{2} \and Truyen Tran\inst{1}} 
 
%\authorrunning{T. Nguyen et al.} 

\institute{
Applied Artificial Intelligence Institute $(\textrm{A}^{2}\textrm{I}^{2})$, Deakin University, Australia \\
\email{\{minh.t.nguyen,dung.nguyen,truyen.tran\}@deakin.edu.au } 
\and Mathematics Department, Ho Chi Minh City University of Education, Vietnam \\ \email{khapt@hcmue.edu.vn}
}
\maketitle 

\global\long\def\ModelSP{\text{SP-PINN}}%
\global\long\def\ModelMP{\text{MP-PINN}}%

\begin{abstract}
Forecasting temporal processes such as virus spreading in epidemics
often requires more than just observed time-series data, especially
at the beginning of a wave when data is limited. Traditional methods
employ mechanistic models like the SIR family, which make strong assumptions
about the underlying spreading process, often represented as a small
set of compact differential equations. Data-driven methods such as
deep neural networks make no such assumptions and can capture the
generative process in more detail, but fail in long-term forecasting
due to data limitations. We propose a new hybrid method called $\ModelMP$
(Multi\textbf{-}Phase Physics-Informed Neural Network) to overcome
the limitations of these two major approaches. $\ModelMP$ instils
the spreading mechanism into a neural network, enabling the mechanism
to update in phases over time, reflecting the dynamics of the epidemics
due to policy interventions. Experiments on COVID-19 waves demonstrate
that $\ModelMP$ achieves superior performance over pure data-driven
or model-driven approaches for both short-term and long-term forecasting.


\end{abstract}

\section{Introduction}

The COVID-19 pandemic has claimed over 7 million lives for just 4
years\footnote{https://www.worldometers.info/coronavirus/coronavirus-death-toll}.
Let us pause for a moment and consider a hard \emph{counterfactual}
question: Could the majority of these lives have been saved if we
had predicted the spread better at the onset of the pandemic and acted
more effectively? In fact, the world did all it could: modelling,
forecasting, implementing lockdowns, developing vaccines, and much
more. In the absence of proper understanding of the viruses' nature
and with limited data available when a wave just started, epidemiologists
had to make assumptions in \emph{model-driven} methods, such as those
in the mechanistic SIR family \cite{kermack1927contribution} or in
detailed agent-based simulations \cite{kerr2021covasim}. When some
data became available, for example, after a month, \emph{data-driven}
methods, as preferred by the data science community, could be employed
to detect trends in the time-series \cite{rodriguez2022data,syed_ziaur_rahman__2023}.
A key open challenge is the complex interplay of evolving interventions,
human factors and technological advances driving the epidemic waves
\cite{burg2023trajectories}.

\begin{figure}
\begin{centering}
\includegraphics[width=0.7\textwidth]{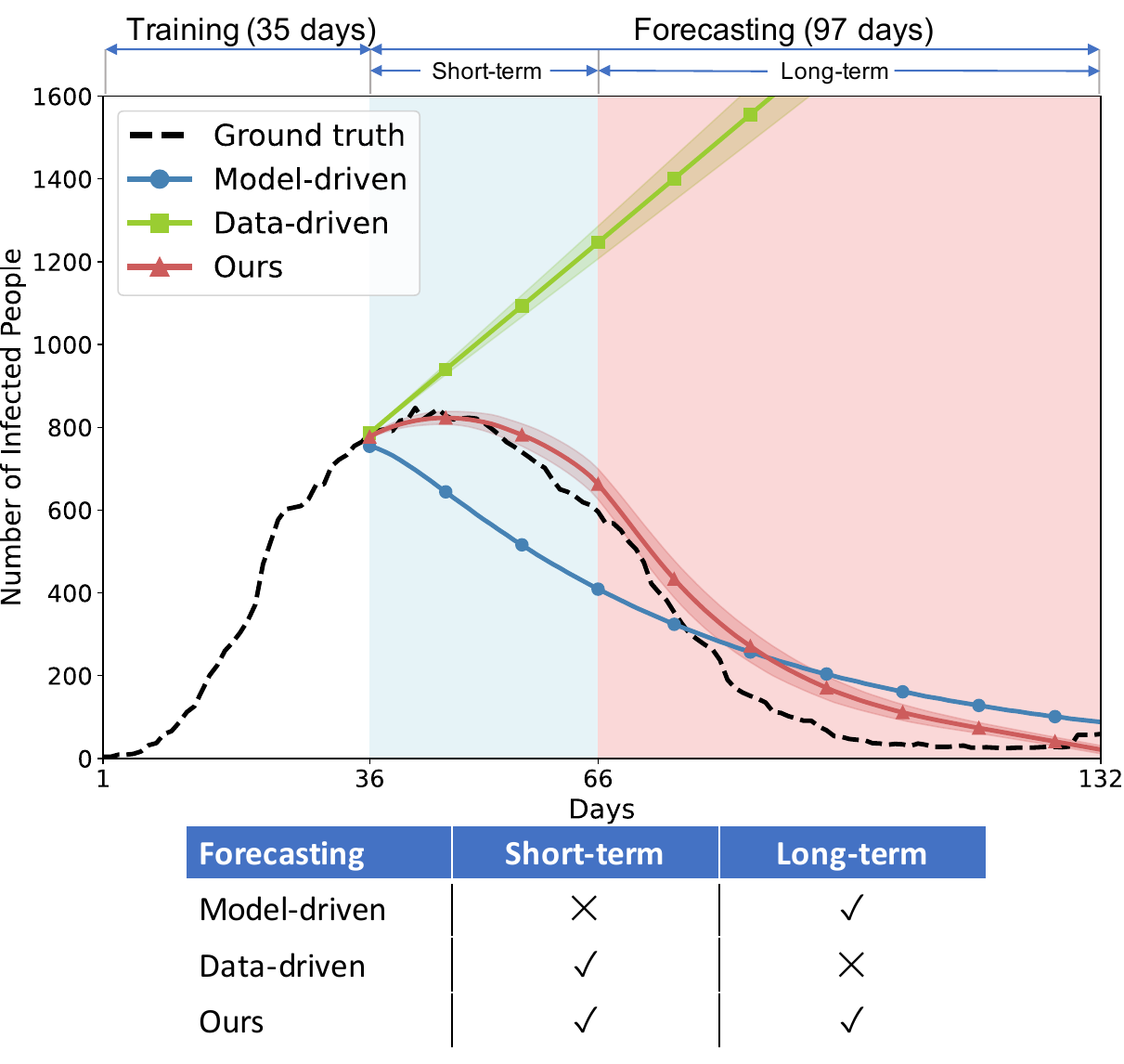}
\par\end{centering}
\caption{A representative case of forecasting COVID-19 at 35 days of the wave.
Our hybrid multi-phase method $\protect\ModelMP$ strikes a balance
between model-driven and data-driven approaches, and hence is more
accurate in both short/long-term forecasting.\label{fig:teaser}\vspace{-5mm}
}
\end{figure}

If anything, the extremely high death toll has profoundly demonstrated
one thing: We have failed to predict the spread of COVID-19 virus
variants. Fig.~\ref{fig:teaser} clearly illustrates this failure.
As seen, model-driven methods such as mechanistic SIR models capture
the overall shape of the wave but are inadequate in reflecting current
data and the changing reality on a daily basis. This might be due
to the rigid and strong assumptions made at the modelling time. Data-driven
methods, such as those using deep neural networks, fit the new evidence
better but fail to capture the long-term underlying mechanisms. Clearly,
a better approach is needed to (a) capture both the short-term and
long-term processes \cite{rahmandad2022enhancing,pagel2022role},
and (b) dynamically calibrate the models in the face of new evidence
\cite{perakis2023covid}.

To this end, we propose $\ModelMP$ (which stands for Multi-Phase
Physics-Informed Neural Network) to overcome these limitations. $\ModelMP$
employs a recent powerful approach known as Physics-Informed Neural
Network (PINN), which trains a neural network to agree with both empirical
data and epidemiological models. However, PINNs alone are not sufficient
to reflect reality: We must account for the complex interplay of evolving
factors driving the epidemic waves, such as changing regulations,
emerging information, and shifting public sentiment, all of which
influence the pandemic's trajectory. This is where $\ModelMP$ comes
in: Instead of assuming a single set of parameters for the entire
wave, we allow the model to vary across \emph{multiple distinct phases},
each represented by a set of SIR parameters. This brings adaptability
into $\ModelMP$.

We demonstrate $\ModelMP$ on COVID-19 data sets collected from COVID-19
data from 21 regions in Italy in the first half of 2020. The results
show that MP-PINN achieves superior performance in both short-term
and long-term forecasting of COVID-19 spread. In particular, $\ModelMP$
outperforms traditional SIR models, pure data-driven approaches (MLP),
and single-phase PINNs. See Fig.~\ref{fig:teaser} for a representative
case demonstrating the efficacy of $\ModelMP$.

\section{Related Works}

\selectlanguage{australian}%

\subsection*{Epidemic Forecasting}

We briefly review the literature in epidemic forecasting most relevant
to our work as the literature has exploded since the COVID-19 outbreak
in late 2019:\vspace{-1mm}

\paragraph*{Model-driven approach}

Compartmental models, such as the Susceptible-Infectious-Recovered
(SIR) model, are foundational in epidemic modelling due to their simplicity
and reliance on mechanistic understanding of disease spread. They
are particularly effective for long-term forecasting because they
incorporate known epidemiological dynamics \cite{kermack1927contribution}.
However, their fixed parameters often lead to less accurate short-term
predictions, as they cannot easily adapt to rapid changes in transmission
dynamics. The have been a plethora of SIR extensions with sophisticated
assumptions such as SIRD \cite{long2021identification}, SEIR \cite{delli2022hybrid}
and SEIRM \cite{rodriguez2023einns}. \vspace{-1mm}

\paragraph*{Data-driven approach}

Machine learning models, particularly deep learning techniques, have
gained prominence for short-term forecasting due to their ability
to capture complex patterns in large datasets \cite{chakraborty2020real}.
Techniques such as Long Short-Term Memory (LSTM) networks excel in
identifying trends and making predictions over short periods. However,
their performance deteriorates over longer horizons due to their lack
of incorporation of epidemiological knowledge, making them less reliable
for long-term predictions \cite{shaman2013real}. In \cite{liu2023epidemiology},
authors studied forecasting influenza outbreaks, e.g. training model
with data in four years to predict in the future outbreaks, hence
the training data is much larger than ours which consists of only
one outbreak.

Most existing data-driven approaches make short-term predictions (less
than a month). For example, the work in \cite{kapoor2020examining}
used two months to train and predict the next one months with prediction
windows are $3/7/14$ days. Likewise, the model in \cite{wang2022causalgnn}
trained on almost ten months and test on one month with prediction
window is $7/14/21/28$ days. In \cite{moon2023reseat} and \cite{sesti2021integrating},
models were trained on $377$ days,  but window of forecasting in
both works is the next $7$ days (observed previous $21$ days). 
In contrast, we train the model only on data collected for \textbf{$35$
}days, and forecast the rest of the outbreak\textbf{ }($97$ days).
Thus our setting is much more challenging and almost impossible to
achieve without utilising epidemic models and prior knowledge.

\subsection*{Physic-Informed Neural Networks (PINNs)}

Recently, PINNs have emerged as a framework to incorporate known physical
rules to train deep neural networks \cite{raissi2019physics}. It
has been demonstrated to be effective in solving forward and inverse
problems involving partial differential equations (PDEs). Since then,
numerous studies have explored the application of PINNs in various
domains, such as fluid dynamics \cite{cai2021physics}, material science
\cite{pun2019physically}, and epidemic modelling \cite{rodriguez2023einns}.
\vspace{-1mm}

\paragraph*{PINNs for epidemic modelling}

PINNs have been used to build hybrid forecasting models, integrating
model-driven and data-driven methods \cite{guo2023comparative,ning2023epi,rodriguez2023einns,shaier2021data}.
The work in \cite{rodriguez2023einns} proposed to regularise the
embeddings from both the time-dependent model which can be regularised
via physical law and the exogenous features extractor which obtain
information from multiple sources for making better predictions. Wang
et al. \cite{wang2020epidemiological} proposed a physics-informed
neural network (PINN) framework for learning the parameters of a COVID-19
compartmental model from observed data. In \cite{gao2021stan}, the
authors used the epidemic model to compute ahead the unobserved data
points then augment the training process directly with the prediction
loss. In \cite{gao2021stan,tang2023enhancing}, although the parameters
of the epidemic model, e.g. the infection rate and the recovery rate,
are generated by a trainable NNs-based module, the potential impact
of data instability on the learned epidemic parameters is not explicitly
addressed. For instance, if case counts change significantly and differ
from historical patterns, it can be challenging for the \cite{gao2021stan,tang2023enhancing}
models to learn valid and stable transmission and recovery rates.
In contrast, our work introduces distinct phases within the SIR model
to capture the long-term dynamics of an outbreak, which may not be
apparent in short-term or noisy data. 

Building on the strengths and limitations of these approaches, our
proposed $\ModelMP$ framework aims to address the gaps in both model-driven
and data-driven methods.\selectlanguage{british}%

\section{Preliminaries }

\subsection{Mechanistic Compartment Models \label{subsec:SIR-model}}

In epidemiology and other fields where the behaviour of large populations
is studied, the compartment models are models in which the population
are divided into a discrete set of qualitatively-distinct states/types/classes/groups,
so-called \emph{compartments}. These models also define the transition
between the compartments.\emph{ }We focus on SIR (Susceptible-Infected-Recovered),
the most popular compartments model used to model the spread of infectious
diseases in epidemiology \cite{kermack1927contribution}\emph{.}
SIR contains three compartments: (1) Susceptible, (2) Infected, (3)
Recovered/removed populations which are denoted as $S(t)$, $I(t)$
and $R(t)$ as a function of time $t$, respectively. At time $t$,
an individual in the population is classified into one of these compartments,
typically transiting from being susceptible to infected and finally
recovered (or removed). Hence, the size of the population ($N$) is
the sum of the number of the susceptible, infectious and recovered
persons, i.e., $N=S(t)+I(t)+R(t)$. Here, we considered a \emph{single
outbreak }SIR model comprises a set of ordinary differential equations
(ODEs) that describe the transitions between three compartments: 

\begin{align}
\frac{dS(t)}{dt} & =-\frac{\beta}{N}S(t)I(t),\label{eq:SIR_dS_dt}\\
\frac{dI(t)}{dt} & =\frac{\beta}{N}S(t)I(t)-\gamma I(t),\label{eq:SIR_dI_dt}\\
\frac{dR(t)}{dt} & =\gamma I(t),\label{eq:SIR_dR_dt}
\end{align}
where parameters of the model $\beta>0$ and $\gamma\in(0,1)$ are
the \emph{infection rate} and the \emph{recovery rate}, respectively.
In real-world and more complex models, these parameters can also vary
over time or depend on different factors such as policy or other properties
of the population. The initial condition of the ODEs are $S(0)>0,I(0)>0$,
and $R(0)\geq0$. 

An important assumption of the SIR model is that all recovered individuals
(in the $R$ group) are completely immune and cannot return to the
susceptible ($S$) or infected ($I$) groups, and that the total population
$N$ remains constant. However, $N$ does not always represent the
entire population, especially in real-world scenarios. For instance,
one assumption of the SIR model \cite{johnson2009mathematical} is
that the population mixes homogeneously, meaning everyone has the
same level of interaction with others. However, during the early stages
of the COVID-19 outbreak in 2020, it was impractical to consider the
entire population as susceptible. Instead, $N$ might only represent
a fraction of the population. Moreover, the population could therefore
be divided into two distinct groups \cite{zhang2021effect}: those
with inherited immunity and those without. These distinctions imply
that the total number of susceptible individuals $S(t)$ may not always
correspond to the entire population $N$, but rather to a specific
portion of it, depending on factors such as inherited immunity or
other epidemiological considerations.

\subsection{Physics-Informed Neural Networks (PINNs) \label{subsec:Physics-Informed-Neural-Networks}}

PINNs \cite{raissi2019physics} are neural networks equipped with
physical constraints of the domain, either through the network architecture
or as a regularisation term during the training process. These physical
laws often involve parameters that need to be estimated. During the
training of a PINN, both the neural network weights and the physical
parameters are optimised to achieve the best fit to the observed data
while simultaneously satisfying the physical constraints.

More formally, given a training dataset $\mathcal{D}$, learning searches
for a neural network $f\in\mathcal{H}$ by solving an optimisation
problem:
\begin{equation}
f^{*}=\min_{f\in\mathcal{H}}\mathcal{L}(f;\mathcal{D})+\lambda\Omega(f),\label{eq:PINN-loss}
\end{equation}
where $\mathcal{L}(f;\mathcal{D})$ is usual data loss function, $\Omega(f)$
is a regularisation term that introduces physical prior knowledge
into the learning process, and $\lambda>0$ is a balancing weight.
When the prior is specified as partial differential equations (PDEs),
the regularisation typically takes the form of the PDE residual loss:
\begin{equation}
\Omega(f)=\frac{1}{L}\sum_{j=1}^{L}\left(\partial f_{\text{NN}}(x_{j})-\partial f{}_{\text{PDE}}(x_{j})\right)^{2}\label{eq:PDE-loss}
\end{equation}
where $\partial f_{\text{NN}}(x_{j})$ denotes the partial derivative
of the neural network evaluated at $x_{j}$, and $\partial f_{\text{PDE}}(x_{j})$
denotes the corresponding partial derivate specified by the PDEs.
The evaluation points $j=1,2,...,L$ are sampled so that the function
$f$ and its derivative are well supported.

\section{Methods}

\begin{figure}
\begin{centering}
\includegraphics[width=0.9\textwidth]{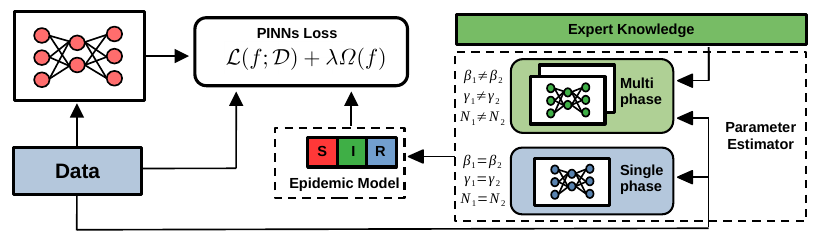}
\par\end{centering}
\caption{Multi-phase Physics-Informed Neural Network ($\protect\ModelMP$)
Framework for Epidemic Forecasting. The framework illustrates the
integration of expert knowledge and data-driven approaches to estimate
parameters in a multi-phase scenario, where key parameters such as
infection rate $\left(\beta\right)$ and recovery rate $\left(\gamma\right)$
vary across phases.\label{fig:MP-PINN-framework}}
\end{figure}

We now present our main contributions of developing PINNs for epdidemic
forecasting. The overall framework is depicted in Fig.~\ref{fig:MP-PINN-framework}.
We start from a single-phase assumption, then advance into the multi-phase
model. Both single-phase and multi-phase models integrate expert knowledge
and a data-driven parameter estimation process.  In the multi-phase
model, however, parameters such as infection rate $\left(\beta\right)$
and recovery rate $\left(\gamma\right)$ vary between phases, enhancing
the model's ability to adapt to complex real-world scenarios.  The
single-phase approach is described in Sec.~\ref{subsec:SP-PINN}
and the multi-phase approach is detailed in Sec.~\ref{subsec:MP-PINN}.

More concretely, using the SIR model described in Sec.~\ref{subsec:SIR-model}
as physics prior, we build a PINN framework with separate neural networks
$f_{\psi_{1}}^{S}(t)$ and $f_{\psi_{2}}^{I}(t)$ where $f_{\psi_{i}}$
that takes $t$ as input to predict the Susceptible $S(t)$ and the
Infected $I(t)$, respectively. For clarity, we present the case where
the input is only the time $t$, but any other features, static or
dynamic can be applicable. Note that we do not need to model the Recovered
$R(t)$ because of the constraint $S(t)+I(t)+R(t)=N$ described in
Sec.~\ref{subsec:SIR-model}. 

\subsection{Single-Phase PINN (SP-PINN) \label{subsec:SP-PINN}}

Consider an outbreak of $T$ days, where we observe the first $T_{0}$
days and forecast the rest of $\left(T-T_{0}\right)$ days. Let $N$
be the size of population who will be infected through the lifetime
of the wave. The PINN's objective function involves an empirical loss
$\mathcal{L}$ and regulariser $\Omega$, as described in Eqs.~(\ref{eq:PINN-loss},\ref{eq:PDE-loss}):
\[
\mathcal{L}=\frac{1}{T_{0}}\stackrel[t=1]{T_{0}}{\sum}\left[\left(f_{\psi_{1}}^{S}\left(t\right)-S\left(t\right)\right)^{2}+\left(f_{\psi_{2}}^{I}\left(t\right)-I\left(t\right)\right)^{2}\right],
\]
where and $S\left(t\right)$ and $I\left(t\right)$ are the ground
truth susceptible and infected populations at observed time $t=1,2,...,T_{0}$.
The physics-informed regulariser has two components: $\Omega=\frac{1}{2}\left(\Omega^{S}+\Omega^{I}\right)$,
where
\begin{align}
\Omega^{S} & =\frac{1}{T}\stackrel[t=1]{T}{\sum}\left(\frac{df_{\psi_{1}}^{S}\left(t\right)}{dt}+\frac{\beta}{N}f_{\psi_{2}}^{I}\left(t\right)f_{\psi_{1}}^{S}\left(t\right)\right)^{2},\label{eq:SIR-loss-S}\\
\Omega^{I} & =\frac{1}{T}\stackrel[t=1]{T}{\sum}\left(\frac{df_{\psi_{2}}^{I}\left(t\right)}{dt}-\frac{\beta}{N}f_{\psi_{2}}^{I}\left(t\right)f_{\psi_{1}}^{S}\left(t\right)+\gamma f_{\psi_{2}}^{I}\left(t\right)\right)^{2}.\label{eq:SIR-loss-I}
\end{align}
These two components $\Omega^{S}$ and $\Omega^{I}$ implement the
residual loss for the ODE of $S(t)$ in Eq.~(\ref{eq:SIR_dS_dt})
and $I(t)$ in Eq.~(\ref{eq:SIR_dI_dt}), respectively. For simplicity,
the balancing weight $\lambda$ in Eq.~(\ref{eq:PINN-loss}) is set
to be $1$. 

\subsubsection{Estimating SIR Parameters}

Recall that in SIR, $\beta$ is the infection rate, $\gamma$ is the
recovery rate. These two parameters and the population size $N$ are
estimated from dataas follows. We first use an ODE solver to get
an initial estimates for the parameters $\beta_{0}$, $\gamma_{0}$,
$N_{0}$ as in classic SIR. We then employ three separate neural networks
$f_{\theta_{1}}^{\beta},f_{\theta_{2}}^{\gamma},f_{\theta_{3}}^{N}$
to refine the estimation for $\beta$, $\gamma$, and $N$ as

\begin{align}
\beta & =\beta_{0}\left(a^{\beta}+b^{\beta}\tanh\left(f_{\theta_{1}}^{\beta}\left(I\left[1:T_{0}\right]\right)\right)\right),\label{eq:indentify_net_beta}\\
\gamma & =\gamma_{0}\left(a^{\gamma}+b^{\gamma}\tanh\left(f_{\theta_{2}}^{\gamma}\left(I\left[1:T_{0}\right]\right)\right)\right),\label{eq:indentify_net_gamma}\\
N & =N_{0}\left(a^{N}+b^{N}\tanh\left(f_{\theta_{3}}^{N}\left(I\left[1:T_{0}\right]\right)\right)\right),\label{eq:indentify_net_N}
\end{align}
where $I[1:T_{0}]$ denotes the observed time-series of the infected
population over time from day $1$ to day $T_{0}$; and $\left\{ a^{\beta},b^{\beta},a^{\gamma},b^{\gamma},a^{N},b^{N}\right\} \geq0$
are scaling hyper-parameters to ensure that $\beta$, $\gamma$, $N$
are within certain ranges of the initial estimates $\beta_{0}$,
$\gamma_{0}$, $N_{0}$, respectively. All neural networks shared
an identical architecture (see Table~\ref{tab:MLP_setting} for more
details). In our experiments with data from Italy (as described in
Sec.~\ref{par:Datasets-and-Pre-processing}), we set $a^{\beta}=a^{\gamma}=a^{N}=1$,
$b^{\beta}=0.6$, $b^{\gamma}=0$ and $b^{N}=0.9$. 

We call this model by $\ModelSP$ (which stands for \emph{Single-Phase
Physics-Informed Neural Network}).

\subsection{Multi-Phase PINN (MP-PINN)\label{subsec:MP-PINN}}

The singe-phase approach assumes that the SIR parameters are unchanged
during the entire epidemic. This might suffice when the epidemic is
well contained and the intervention does not change over time. This
is unrealistic in the case of COVID-19. We found that the $\ModelSP$
demonstrated good performance for short-term forecasting, it struggled
to capture the evolving dynamics of longer-term outbreaks (see Table~\ref{tab:MLP_setting}
for more details). This limitation motivated the development of our
multi-phase PINN, or $\ModelMP$, described below. 

A complex epidemic outbreak at the scale of COVID-19 is typically
handled by a sequence of interventional policies (e.g., social distancing,
restrictions, lock-downs, and rolling out of vaccines) to adaptively
control the virus spreading and minimise the damage in the public
health and economy of a region. Thus, it is reasonable to assume that
there are $M$ distinct phases, one following another, in the entire
life of a wave, each of which can be characterised by a parameter
set $\left(\beta_{p},\gamma_{p}\right)_{p=1}^{M}$ in the SIR model.
Thus Eqs.~(\ref{eq:SIR-loss-S},\ref{eq:SIR-loss-I}) are extended
to:
\begin{align}
\Omega_{\text{multi}}^{S} & =\frac{1}{T}\sum_{p=1}^{M}\sum_{t=\tau_{p}}^{\tau_{p+1}}\left(\frac{df_{\psi_{1}}^{S}(t)}{dt}+\frac{\beta_{p}}{N}f_{\psi_{2}}^{I}(t)f_{\psi_{1}}^{S}(t)\right)^{2},\label{eq:SIR-loss-S-1}\\
\Omega_{\text{multi}}^{I} & =\frac{1}{T}\sum_{p=1}^{M}\sum_{t=\tau_{p}}^{\tau_{p+1}}\left(\frac{df_{\psi_{2}}^{I}(t)}{dt}-\frac{\beta_{p}}{N}f_{\psi_{2}}^{I}(t)f_{\psi_{1}}^{S}(t)+\gamma_{p}f_{\psi_{2}}^{I}(t)\right)^{2},\label{eq:SIR-loss-I-1}
\end{align}
where $\tau_{p}$ denotes the starting date of phase $p$ where $p=1,2,...,M$
with $\tau_{1}=1$.

The key for this model is therefore the set of dates $\left\{ \tau_{p}\right\} _{p=2}^{M}$
where new phases are in effect. In our experiments, for $M=2$, we
set $\tau_{2}=T_{0}+30$ for simplicity. 

\subsubsection{Estimating SIR Parameters}

We follow the same estimation procedure as in the case of single phase,
that is, we first estimate $\beta_{0}$, $\gamma_{0}$ and $N_{0}$
using an ODE solver, then refine them for each phase. The estimations
of parameters in phases $\left\{ \left(\beta_{p},\gamma_{p}\right)\right\} _{p=1\dots M}$
follow Eqs.~(\ref{eq:indentify_net_beta},\ref{eq:indentify_net_gamma},\ref{eq:indentify_net_N})
with different constraints $\left\{ \left(a_{p}^{\beta},b_{p}^{\beta},a_{p}^{\gamma},b_{p}^{\gamma},a_{p}^{N},b_{p}^{N}\right)\right\} _{p=1\dots M}$

\begin{align}
\beta_{p} & =\beta_{0}\left(a_{p}^{\beta}+b_{p}^{\beta}\tanh\left(f_{\theta_{1p}}^{\beta}\left(I\left[1:T_{0}\right]\right)\right)\right),\label{eq:indentify_net_beta_p}\\
\gamma_{p} & =\gamma_{0}\left(a_{p}^{\gamma}+b_{p}^{\gamma}\tanh\left(f_{\theta_{2p}}^{\gamma}\left(I\left[1:T_{0}\right]\right)\right)\right),\label{eq:indentify_net_gamma_p}\\
N_{p} & =N_{0}\left(a_{p}^{N}+b_{p}^{N}\tanh\left(f_{\theta_{3p}}^{N}\left(I\left[1:T_{0}\right]\right)\right)\right).\label{eq:indentify_net_N_p}
\end{align}

In COVID-19 forecasting where we consider two phases short-term and
long-term prediction, i.e. $M=2$, the constraints can be determined
as follows. Since the population is assumed to remain unchanged, we
use the same estimation of $N_{p}$ for all phases $p=1,\dots,M$.
Our framework allows experts to specify constraints $\left\{ \left(a_{p}^{\beta},b_{p}^{\beta},a_{p}^{\gamma},b_{p}^{\gamma},a_{p}^{N},b_{p}^{N}\right)\right\} _{p=1\dots M}$
based on understanding about the epidemic, modelling assumptions and
real-world scenarios. The experts can rely on trending in recovery
rate and infection rate due to the policies from government and changes
in behaviour of the population to set $a_{p}^{\beta},a_{p}^{\gamma},$
and $a_{p}^{N}$. Furthermore, the experts might be more uncertain
about parameters in phases that data is unavailable than in the phase
that he/she observes the data, therefore, $b_{1}^{\beta}<b_{2}^{\beta}$
and $b_{1}^{\gamma}<b_{2}^{\gamma}$. In our experiments with data
from Italy (as described in Sec.~\ref{par:Datasets-and-Pre-processing}),
we set $a_{1}^{\beta}=a_{2}^{\beta}=a_{1}^{\gamma}=1$, $a_{2}^{\gamma}=2$,
$b_{1}^{\beta}=0.6<b_{2}^{\beta}=0.999$ and $b_{1}^{\gamma}=0.01<b_{2}^{\gamma}=1.0$.

We call this model by $\ModelMP$ (which stands for \emph{Multi-Phase
Physics-Informed Neural Network}).

\section{Experiments}

\subsection{Settings }

\begin{figure}
\begin{centering}
\includegraphics[width=0.8\textwidth]{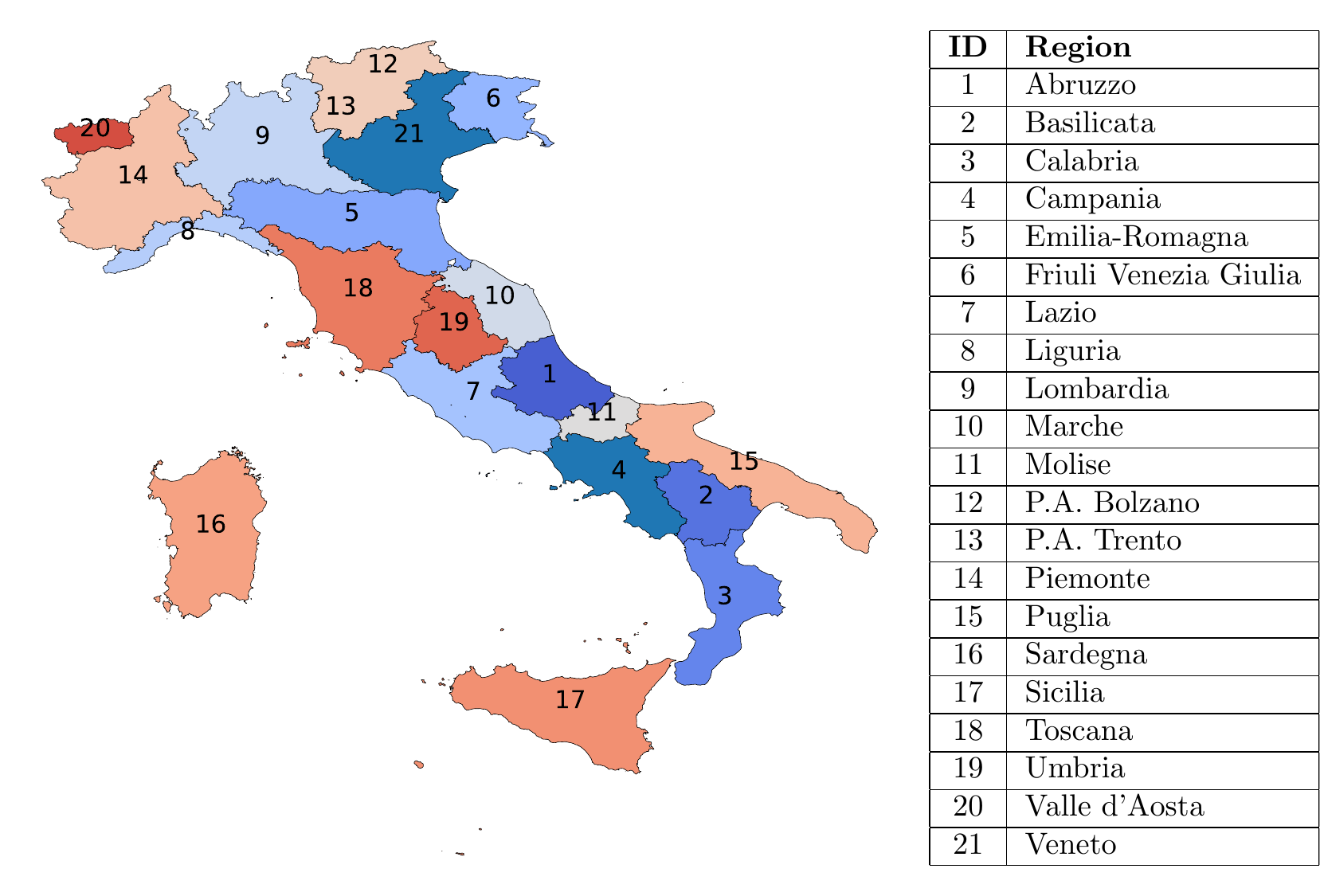}
\par\end{centering}
\caption{The map of Italy with 21 regions created using Geopandas \cite{geopandas}
and data from the GeoJSON file \cite{openpolis2024}.\label{fig:The-regional-map}}
\end{figure}

\subsubsection{Datasets and Pre-processing \label{par:Datasets-and-Pre-processing}}

We use the COVID-19 data\footnote{https://github.com/pcm-dpc/COVID-19}
collected in $21$ regions of Italy \cite{morettini2020covid} from
February $24^{\text{th}}$ 2020 to August $30^{\text{th}}$ 2020.
Fig.~\ref{fig:The-regional-map} shows the map and the region list.
For our purpose of studying a forecasting task in a single outbreak,
we extracted data in $132$ consecutive days, which is from March
$5^{\text{th}}$ 2020 to July $14^{\text{th}}$ 2020. In the first
day, the number of infected persons are in all regions. We used the
total amount of current positive cases (TPC), in the data as the number
of infected persons $I(t)$ in each region. The number of recovered
persons $R(t)$ in each region is calculated by the sum of the recovered
and death cases. The total population data for $2020$$\left(N_{2020}\right)$
was obtained from the ISTAT database\footnote{http://dati.istat.it}.

For our forecasting task, we divided this data into two sets: (1)
a training set, which contains the data in $35$ days from the beginning
of the epidemic and (2) a test set, which contains data in the last
$97$ days of the extracted single outbreak.

Our real-world data analysis utilises COVID-19 infection, recovery,
and mortality rates collected from multiple regions, with daily updates
on case counts distributed across different postal areas. This dataset
provides a comprehensive view of the pandemic's evolution over time,
allowing for a detailed evaluation of $\ModelSP$ and $\ModelMP$
in predicting disease trends under varying conditions.

\subsubsection{Baselines}

We implement SIR and Multilayer Perceptron (MLP) for two representative
baselines as SIR represents the model-driven approach, and MLP represents
the data-driven approach.

\paragraph{SIR}

The framework begins with the initialisation of the population size
($N$) and the initial conditions for the susceptible ($S_{0}$),
infectious ($I_{0}$), and recovered ($R_{0}$) compartments. These
values serve as inputs to a standard Ordinary Differential Equation
(ODE) solver. In this study, the \texttt{\small{}scipy.integrate.solve\_ivp}
solver from the SciPy package was utilised, which generates estimates
of the infectious $\left(\widehat{I}\right)$ and recovered $\left(\widehat{R}\right)$
individuals over time. Subsequently, the observed data for the infectious
($I$) and recovered ($R$) individuals are compared against the model
estimates, and this comparison is quantified using a loss function,
expressed as $\mathcal{L}(I,\hat{I})+\mathcal{L}(R,\hat{R}).$ This
loss function is then minimised through an optimisation process using
the Nelder-Mead optimiser, resulting in estimates for the infection
rate $\widehat{\beta}$, recovery rate $\widehat{\gamma}$, and the
population size $\widehat{N}$. These optimised parameters are used
iteratively in the ODE solver, refining the estimates of $\hat{I}$
and $\hat{R}$ until the model converges. This iterative process provides
a solution to the SIR model that best fits the observed data.

\subsubsection{Metric}

We compute the \emph{symmetric Mean Absolute Percentage Error} (sMAPE)
for each region $j$ as follows

\[
m_{sMAPE}^{j}=\frac{1}{K}\sum_{t=t_{N}}^{t_{N}+K}\frac{\left|y(t)-\hat{y}(t)\right|}{(y(t)+\hat{y}(t)+\epsilon)/2},
\]
where $\epsilon=10^{-32}$ is added to prevent division by zero when
actual values $y(t)$ or predicted values $\hat{y}(t)$ are equal
to $0$. We opt for sMAPE over MAPE because sMAPE addresses the issue
of asymmetry in MAPE. In MAPE, the error can be disproportionately
large when the actual value $y(t)$ is small, whereas sMAPE normalises
this by considering both the predicted and actual values, leading
to a more balanced and reliable metric, especially in cases where
values may approach zero. The final metric is computed as the average
of over $21$ regions, i.e. $m_{sMAPE}^{total}=\frac{1}{21}\sum_{j=1}^{21}m_{sMAPE}^{j}$.
The better forecaster will give the smaller value of $m_{sMAPE}^{total}$.
\vspace{-5mm}

\begin{table}[t]
\caption{Multi-layer Perceptron (MLP) configurations used for estimating SIR
parameters and the susceptible (S) and infected (I) populations. The
table details the dimensionality, number of layers, and activation
functions applied in each MLP type.\label{tab:MLP_setting}\vspace{2mm}
}

\centering{}%
\begin{tabular}{|l|c|c|c|c|}
\hline 
MLP Type & Layer & Size & \# Layers & Activation\tabularnewline
\hline 
\hline 
\multirow{3}{*}{Type 1: SIR Parameters} & Input & $35$ & $1$ & Tanh\tabularnewline
\cline{2-5} \cline{3-5} \cline{4-5} \cline{5-5} 
 & Middle & $64$ & $2$ & Tanh\tabularnewline
\cline{2-5} \cline{3-5} \cline{4-5} \cline{5-5} 
 & Output & $1$ & $1$ & -\tabularnewline
\hline 
\multirow{3}{*}{Type 2: Susceptible (S) \& Infected (I)} & Input & $1$ & $1$ & CELU\tabularnewline
\cline{2-5} \cline{3-5} \cline{4-5} \cline{5-5} 
 & Middle & $32$ & $4$ & CELU\tabularnewline
\cline{2-5} \cline{3-5} \cline{4-5} \cline{5-5} 
 & Output & $1$ & $1$ & -\tabularnewline
\hline 
\end{tabular}
\end{table}

\begin{figure}
\begin{centering}
\includegraphics[width=0.9\textwidth]{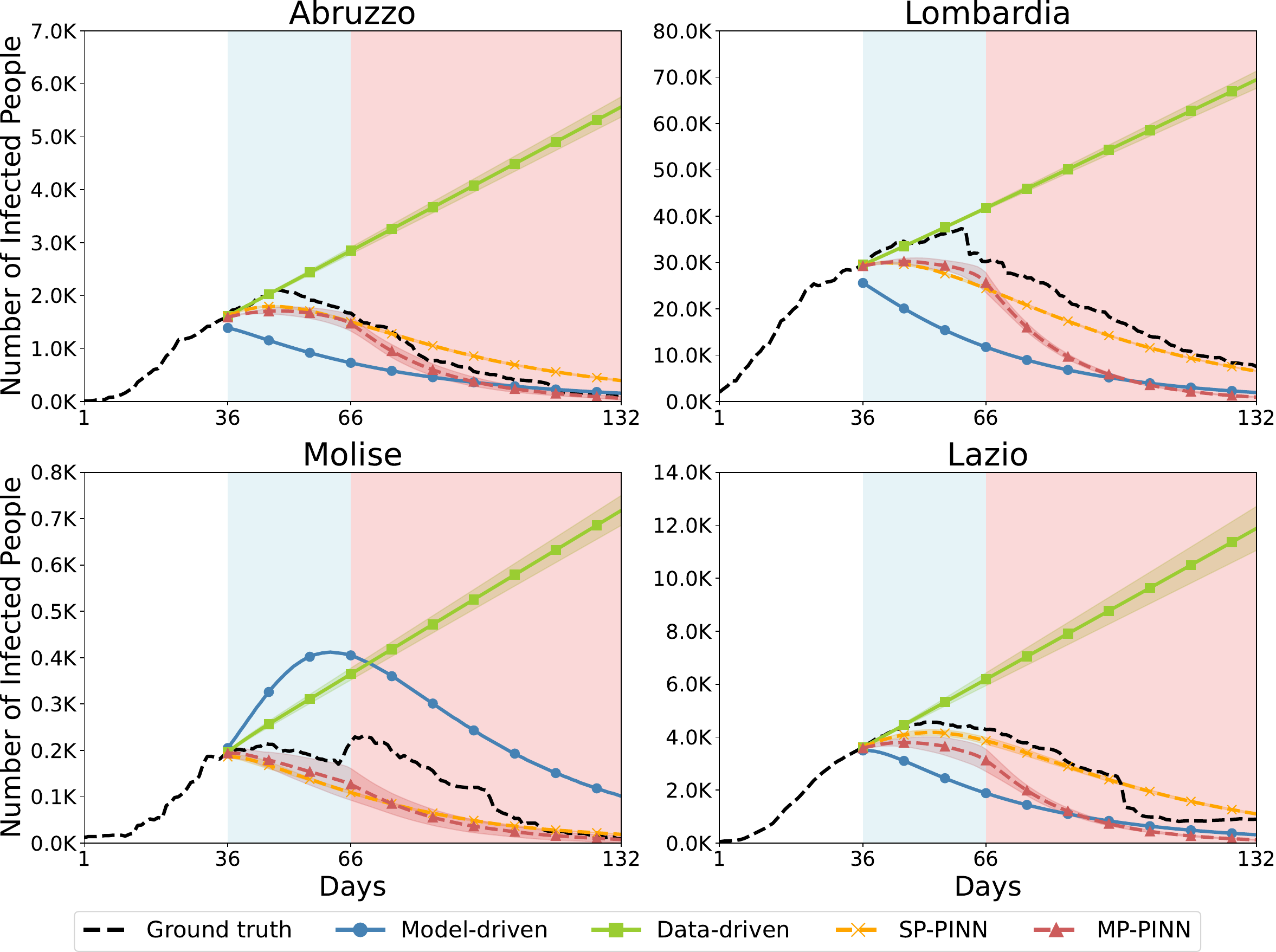}
\par\end{centering}
\caption{Forecasting examples of short-term (next 30 days) and long-term (beyond
30 days) demonstrating the superior performance of PINNs like $\protect\ModelSP$
and $\protect\ModelMP$ compared to pure model-driven (SIR) and data-driven
(MLP) methods. The background colours indicate the training period
(white), short-term forecasting horizon (blue), and long-term forecasting
horizon (red).\label{fig:Effectivenes-of-hybrid-models} }
\end{figure}

\begin{figure}
\begin{centering}
\includegraphics[width=0.9\textwidth]{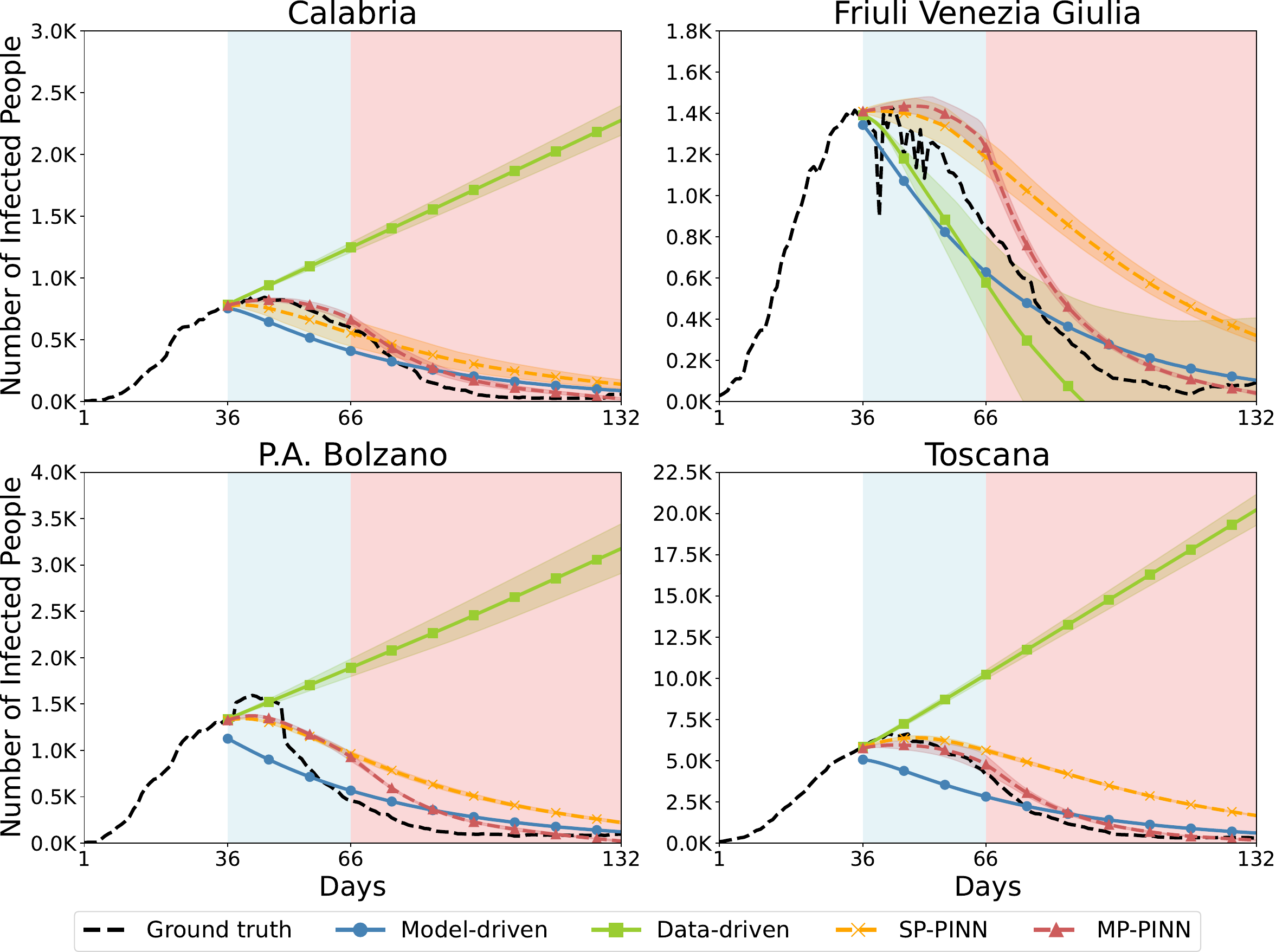}
\par\end{centering}
\caption{Forecasting examples showcasing the importance of multi-phase modelling
in $\protect\ModelMP$. When the dynamics changes, single models are
inadequate. The background colours indicate the training period (white),
short-term forecasting horizon (blue), and long-term forecasting horizon
(red).\label{fig:Effectiveness-of-MP-PINN}}
\end{figure}

\subsubsection{Experiment Setup }

In absence of true timing of phrases, we configure a two-phase $\ModelMP$,
where the next phase starts at $\tau_{2}=T_{0}+30$ (the first phase/short-term
prediction includes one month after the end of training data). For
training models, we employ the Adam optimiser, with a learning rate
of $3\times10^{-4}$ and a total of 30,000 epochs. We utilise two
distinct types of Multi-layer Perceptrons (MLPs) to estimate the parameters
of the SIR model and the state variables, as described in Table.~\ref{tab:MLP_setting}.
In this setup, each region was trained independently, with a separate
MLP for each parameter (population size $N$, infection rate $\beta$,
recovery rate $\gamma$, susceptible persons $S$, and infected persons
$I$).

\subsection{Results}

Forecasting cases are shown in Figs.~\ref{fig:Effectivenes-of-hybrid-models}
and \ref{fig:Effectiveness-of-MP-PINN}. As shown in Fig\@.~\ref{fig:Effectivenes-of-hybrid-models},
single phase PINNs (i.e., the $\ModelSP$) can be very competitive
in short-term forecasting, and reliable in long-term forecasting.
Fig.~\ref{fig:Effectiveness-of-MP-PINN} demonstrates that multi-phase
PINNs (i.e., the $\ModelMP$) are superior when the multi-phases hypothesis
holds.

\begin{figure}
\begin{centering}
\includegraphics[width=0.8\textwidth]{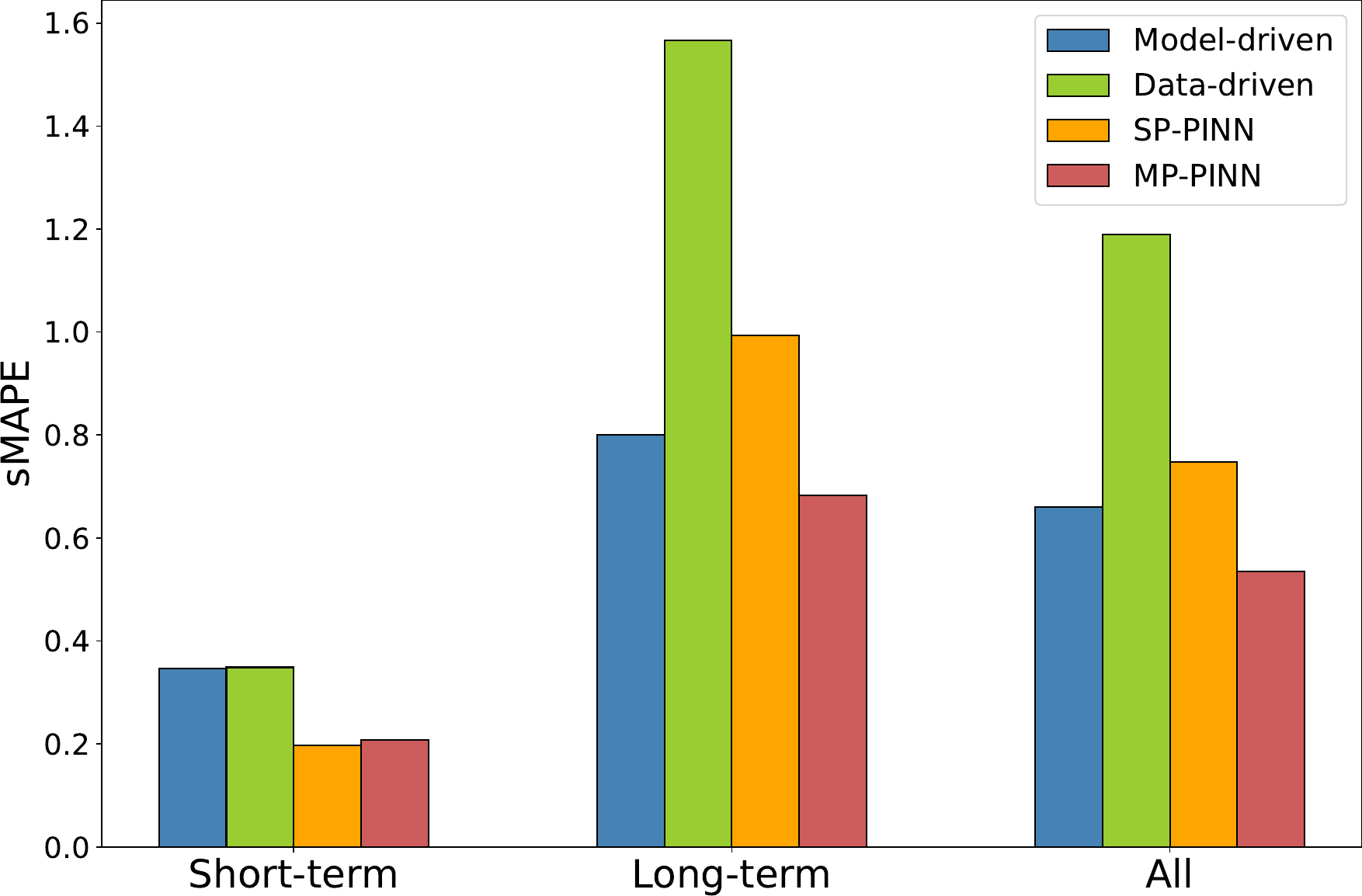}
\par\end{centering}
\caption{sMAPE averaged over all regions at different forecasting horizons:
Short-term (next 30 days), long-term (beyond 30 days), and for the
entire remaining time of the outbreak. Compared methods are: SIR (model-driven),
MLP (data-driven), single phase ($\protect\ModelSP$) and multi-phase
($\protect\ModelMP$). \label{fig:Comparison-of-Mean-sMAPE}}
\end{figure}

The overall performance of methods over all regions are summarised
in Fig.~\ref{fig:Comparison-of-Mean-sMAPE} and Table~\ref{tab:sMAPE-all-regions}.
As shown, the pure data-driven method, MLP, performs relatively well
on the short-term forecast, but fails miserable on the long-term.
This is understandable, as MLP tends to capture the local trend, and
totally ignorant of the nature of the virus spreading dynamics. The
pure model-driven method, SIR, performs generally well overall, reflecting
a nearly century of accumulative knowledge about the virus dynamics. 

The results also demonstrate that the two-phase model $\ModelMP$
effectively captures the two-phase nature of the outbreak. The model's
ability to leverage prior knowledge about the potential ranges of
$\beta$ and $\gamma$ in the second phase significantly improves
its long-term forecasting accuracy. This suggests that incorporating
expert insights can guide the model's learning and enhance its predictive
capabilities, particularly when data or knowledge for the later phases
is limited or uncertain. The success of the multi-phase approach underscores
the potential of $\ModelMP$ to dynamically integrate domain expertise
with data-driven learning, leading to more accurate epidemic forecasts.

The results illustrate that PINNs can learn to shape the forecasting
in two phases as well when we have this prior knowledge (also a strong
prior although it does not use any data points in the future). In
this data, it is still difficult to estimate $\beta$ and $\gamma$
for the later phase without any prior knowledge, but having suitable
prior knowledge, PINNs can learn well. 

\begin{table}
\caption{sMAPE averaged over all regions at different forecasting period: Short-term
(next 30 days), long-term (beyond 30 days), and for the entire remaining
time of the outbreak. The smaller the better.\label{tab:sMAPE-all-regions}}

\centering{}%
\begin{tabular}{cccc}
\toprule 
Method & Next 30 days & Beyond 30 days & All days\tabularnewline
\midrule
SIR & 0.346 & 0.800 & 0.660\tabularnewline
MLP & 0.349 & 1.566 & 1.190\tabularnewline
\midrule
$\ModelSP$ & \textbf{0.197} & 0.993 & 0.747\tabularnewline
$\ModelMP$ & 0.208 & \textbf{0.683} & \textbf{0.536}\tabularnewline
\bottomrule
\end{tabular}
\end{table}

\vspace{-5mm}

\section{Conclusions}

\subsection*{\vspace{-5mm}
}

In this work, we introduced $\ModelMP$ (Multi-Phase Physics-Informed
Neural Network), a novel approach to epidemic forecasting that addresses
the limitations of both traditional model-driven and data-driven methods.
By integrating the mechanistic understanding of SIR models with the
flexibility of neural networks and allowing for multiple SIR parameters
across phases, $\ModelMP$ achieves superior performance in both short-term
and long-term forecasting of COVID-19 spread. Our experiments on COVID-19
data from 21 regions in Italy demonstrate that $\ModelMP$ outperforms
traditional SIR models, pure data-driven approaches (MLP), and single-phase
PINNs. The ability to capture evolving dynamics through multiple phases
proves crucial in reflecting the impact of changing interventions
and public behaviours throughout the course of the epidemic. $\ModelMP$'s
success highlights the potential of hybrid approaches that combine
domain knowledge with data-driven learning. By allowing for the incorporation
of expert insights and prior knowledge about parameter ranges, our
method provides a flexible framework that can adapt to the complex,
evolving nature of real-world epidemics. Future work could explore
the automatic detection of phase transition points and the incorporation
of additional epidemiological factors. Finally, we emphasise that
the $\ModelMP$ framework is generally applicable to any outbreaks
where underlying dynamics may shift over time.

\bibliographystyle{plain}
\bibliography{main}

\end{document}